\documentclass[letterpaper]{article} 
\usepackage{aaai25}  
\usepackage{times}  
\usepackage{helvet}  
\usepackage{courier}  
\usepackage[hyphens]{url}  
\usepackage{graphicx} 
\usepackage{natbib}  
\usepackage{caption} 
\usepackage{amssymb}
\usepackage{algorithm}
\usepackage{algorithmic}
\usepackage{url}
\usepackage{booktabs}
\usepackage{amsmath}
\usepackage{newfloat}
\usepackage{listings}
\DeclareCaptionStyle{ruled}{labelfont=normalfont,labelsep=colon,strut=off} 
\floatstyle{ruled}
\newfloat{listing}{tb}{lst}{}
\floatname{listing}{Listing}
\title{MPDS: A Movie Posters Dataset for Image Generation with Diffusion Model}
\author{
    \textsuperscript{\rm 1}Meng Xu, \textsuperscript{\rm 1}Tong Zhang, \textsuperscript{\rm 1}Fuyun Wang, \textsuperscript{\rm 1}Yi Lei, \textsuperscript{\rm 2}Xin Liu, \textsuperscript{\rm 1}Zhen Cui\\
}
\affiliations{
	\textsuperscript{\rm 1}
	Nanjing University of Science and Technology, Nanjing, China. \\
	\textsuperscript{\rm 2}SeetaCloud, Nanjing, China. \\
	\{xumeng, tong.zhang, fyw271828, leiyi\_010760, zhen.cui\} @njust.edu.cn\\
	\{xin.liu\}@seetacloud.com
   
}

\usepackage{bibentry}

\begin{document}

\maketitle

\begin{abstract}
Movie posters are vital for captivating audiences, conveying themes, and driving market competition in the film industry. While traditional designs are laborious, intelligent generation technology offers efficiency gains and design enhancements. Despite exciting progress in image generation, current models often fall short in producing satisfactory poster results. The primary issue lies in the absence of specialized poster datasets for targeted model training. In this work, we propose a Movie Posters DataSet (MPDS), tailored for text-to-image generation models to revolutionize poster production. As dedicated to posters, MPDS stands out as the first image-text pair dataset to our knowledge, composing of 373k+ image-text pairs and 8k+ actor images (covering 4k+ actors). Detailed poster descriptions, such as movie titles, genres, casts, and synopses, are meticulously organized and standardized based on public movie synopsis, also named movie-synopsis prompt. To bolster poster descriptions as well as reduce differences from movie synopsis, further, we leverage a large-scale vision-language model to automatically produce vision-perceptive prompts for each poster, then perform manual rectification and integration with movie-synopsis prompt. In addition, we introduce a prompt of poster captions to exhibit text elements in posters like actor names and movie titles. For movie poster generation, we develop a multi-condition diffusion framework that takes poster prompt, poster caption, and actor image (for personalization) as inputs, yielding excellent results through the learning of a diffusion model. Experiments demonstrate the valuable role of our proposed MPDS dataset in advancing personalized movie poster generation. MPDS is available at https://anonymous.4open.science/r/MPDS-373k-BD3B.
\end{abstract}

\section{Introduction}
\label{introduction}
Movie posters play an indispensable role in the promotion of films, attracting more audiences, enhancing visibility, conveying themes and emotions, and influencing market competition. The traditional manual design processes are resource-intensive and time-consuming. Recently, the emergence of diffusion-based image generation models~\cite{ho2020denoising,song2020denoising,rombach2022high,saharia2022photorealistic,balaji2022ediff,sohl2015deep,yang2023diffusion} has revolutionized the creative landscape, enabling the production of captivating visual content. Leveraging these advanced models to streamline the creation of movie posters can lead to enhanced efficiency and creativity in the era of intelligent industry.

Currently, there is a notable absence of specialized efforts focused on the paradigm of text-to-image generation for movie posters, despite the proliferation of text-to-image diffusion models~\cite{rombach2022high,saharia2022photorealistic,balaji2022ediff,sohl2015deep,yang2023diffusion} originating from broader tasks and datasets~\cite{schuhmann2021laion,schuhmann2022laion,changpinyo2021conceptual,lin2014microsoft,thomee2016yfcc100m,sharma2018conceptual}. 
However, these models frequently fall short in producing satisfactory movie posters due to the lack of movie poster-specific data in their training datasets. 
For instance, the utilization of the LAION-5B dataset~\cite{schuhmann2022laion} to train Stable Diffusion~\cite{rombach2022high}, whereby image sources were sourced from web images and alt text attributes, led to inadequacies as there are not sufficient appropriate pairs of movie posters and text descriptions.
In other word, those prompt-image data across various domains exhibit disparities, with poster-specific data possessing unique characteristics distinct from conventional cases. Models lacking training on poster-specific data struggle to comprehend requisite text prompts, resulting in generated content that deviates from the intended target and diverges from authentic poster aesthetics. As shown in Fig.~\ref{generation5} in subsequent experiments, the conventional text-to-image generation outcomes appear less refined and professional. 
Therefore, a tailored image-text pairs dataset specifically designed for poster generation is under urgent need.

In this work, we propose a Movie Posters DataSet (MPDS), specifically crafted for diffusion models to facilitate movie poster generation. 
To the best of our knowledge, MPDS marks the pioneering image-text pairs dataset tailored specifically for posters. Leveraging data acquisition from some movie websites~\footnote{The main one: www.imdb.com, a preeminent and authoritative repository of movie, TV, and celebrity content.}, we have conducted a substantial collection of movie posters alongside pertinent information, over the span of six months. The curated MPDS totally comprises 373k+ pairs of posters and associative prompts, and 8k+ actor images covering 4k+ actors, laying the foundation for innovative advancements in poster generation research.

Throughout the annotation process, we incorporate text descriptions reflecting both movie synopses and perceptual visual aspects of the posters, i.e., creating movie-synopsis prompts and vision-perceptive prompts.
Given the quantity of hundreds of thousands of posters, relying solely on manual annotation would have posed substantial costs and risks to the accuracy and usability of the annotations. To tackle this challenge, we introduce a semi-manual annotation strategy. Initially, we employ Blip2~\cite{li2023blip}  for the preliminary annotation of the posters, followed by meticulous manual refinement. Following this hybrid annotation approach for producing vision-perceptive prompts, the resulting image-text pairs demonstrate a high degree of accuracy and adequacy, effectively serving for automatic movie poster generation. Besides, we introduce a prompt of poster captions to implant text elements into poster like actor name and movie title. 

To conduct benchmark experiments, we develop a multi-condition diffusion framework that utilizes inputs such as poster prompts, captions, and actor images for personalization. In contrast to those traditional diffusion models trained on conventional datasets, we show the intrinsic value of our innovative MPDS dataset and the effectiveness of our poster-specific diffusion model. It demonstrates the significance of leveraging tailored datasets and specialized models for pushing the boundaries in poster generation research. In summary, our contributions are three folds: i) establish an image-text paired dataset tailored for movie poster generation, bridging an existing gap in this domain; ii) develop a multi-condition diffusion model as a base model for poster generation; iii) conduct benchmark experiments to illustrate the value of the dataset. 
\section{Related Work}
\label{relatedwork}

\paragraph{Image-Text Datasets}
Previous efforts in image-text dataset creation, exemplified by MS-COCO~\cite{lin2014microsoft} and Visual Genome~\cite{krishna2017visual}, involved the curation of image and region labels through human annotation.  While this approach yielded high-quality labels, it constrained the datasets' scale to 0.33 and 5 million examples, respectively.
Most public datasets with millions or billions of image-text paired data, such as LAION-400M~\cite{schuhmann2021laion}, LAION-5B~\cite{schuhmann2022laion}, CC-12M~\cite{changpinyo2021conceptual}, YFCC100M~\cite{thomee2016yfcc100m}, are sourced from the internet. 
The LAION-5B dataset, a widely utilized resource for image-text pairs, played a crucial role in training stable-diffusion model.  This dataset is sourced from Common Crawl~\cite{commoncrawl}, a non-profit organization that regularly releases HTML data from billions of public websites.  
The LAION-5B dataset meticulously extracts HTML IMG tags with alt-text attributes to assemble its extensive collection.
However, an analysis of HTML from movie information websites indicates that the alt-text for poster IMG tags often lacks a literal description. As a result, datasets such as LAION, which employ similar processing methodologies, tend to omit poster-related information. Our study bridges this gap by creating precise literal descriptions for movie posters, thereby enabling the inclusion of poster information in the training of text-to-image models.

\paragraph{Vision-Language Models}

VLM~\cite{zhou2022learning,zhu2023minigpt,zhou2022conditional,radford2021learning,ramesh2021zero,li2023blip} seek to enhance performance in downstream vision and language tasks by pretraining on extensive image-text pairs.  Given the high cost of obtaining human-annotated texts, most approaches~\cite{radford2021learning,li2023blip} rely on image and alt-text pairs sourced from the web.  
CLIP~\cite{radford2021learning}, a model designed to assess the relationship between text and images, has sparked a renewed wave of research interest. By linking text and images, CLIP facilitates the development of transferable visual models. Despite the application of simple filters, web-derived texts often contain significant noise, rendering them suboptimal for vision-language learning. 
BLIP2~\cite{li2023blip} introduces a new training method for unifying multiple vision-language tasks into a joint framework (such as mage-text retrieval and image captioning) that allows for more efficient use of web datasets.
In this study, we utilize the superior performance of BLIP2 to generate textual descriptions for movie posters.

\paragraph{Text-to-Image Model}
The diffusion model ~\cite{nichol2021glide,dhariwal2021diffusion,li2024blip,rombach2022high} has become the preferred choice in the field of text-to-image generation due to its ability to produce stable, high-quality results. 
The latent diffusion model~\cite{rombach2022high} uses an autoencoder to compress the image into the latent space, generating high-resolution images and significantly improving computational efficiency.
Recent large-scale text-to-image models, such as Imagen~\cite{saharia2022photorealistic} DALL-E2~\cite{ramesh2022hierarchical}, Parti~\cite{yu2022scaling}, CogView2~\cite{ding2022cogview2}, and Stable Diffusion~\cite{rombach2022high}, have demonstrated unprecedented capabilities in generating high-quality, high-resolution images.
However, these models lack fine-grained control over the generated images, making it difficult to maintain the consistency of the target identity in the synthesized images.
Some subject-driven text-to-image generation models~\cite{chen2024textdiffuser,gal2022image,ruiz2023dreambooth,zhang2024brush,chen2024subject} maintain the consistency of target identity in the synthesized image, and improve the model's lack of fine-grained control over the generated image.
Textual Inversion~\cite{gal2022image} learns to represent a user-provided concept through new "words" in the embedding space of a frozen text-to-image model. 
In this paper, we conducted some experiments on Stable Diffusion~\cite{rombach2022high}, Textual Inversion~\cite{gal2022image}, DreamBooth~\cite{ruiz2023dreambooth}, Brush Your Text~\cite{zhang2024brush}.

\section{MPDS Dataset}
\label{headings}

\subsection{Initial Data Collection}

Some Image-Text Datasets, such as LAION-400M, obtain HTML-like data from Common Crawl because they aim to collect a wide variety of image-text pairs. However, our goal is specifically to acquire movie posters. We reviewed numerous movie information websites, and finally, we chose to obtain the initial data from the Internet Movie Database (IMDB\footnote{https://www.imdb.com}), an online database of information related to film, television series, and podcasts. We used a sequential strategy to obtain movie details. We attempted to access URLs numbered from 1 to 1,250,000, among which 400k URLs were valid. We extracted the movie title, genre, synopses, main cast, and URLs for downloading the movie poster from the movie detail pages. Additionally, we attempted to access URLs numbered from 1 to 60,000 for actor details, among which 5k URLs were valid. From the actor detail pages, we extracted the actor's name and the URL for downloading the actor's image.

\subsection{Data Processing}
\begin{figure}[h]
	\centering
	\includegraphics[width=1\linewidth]{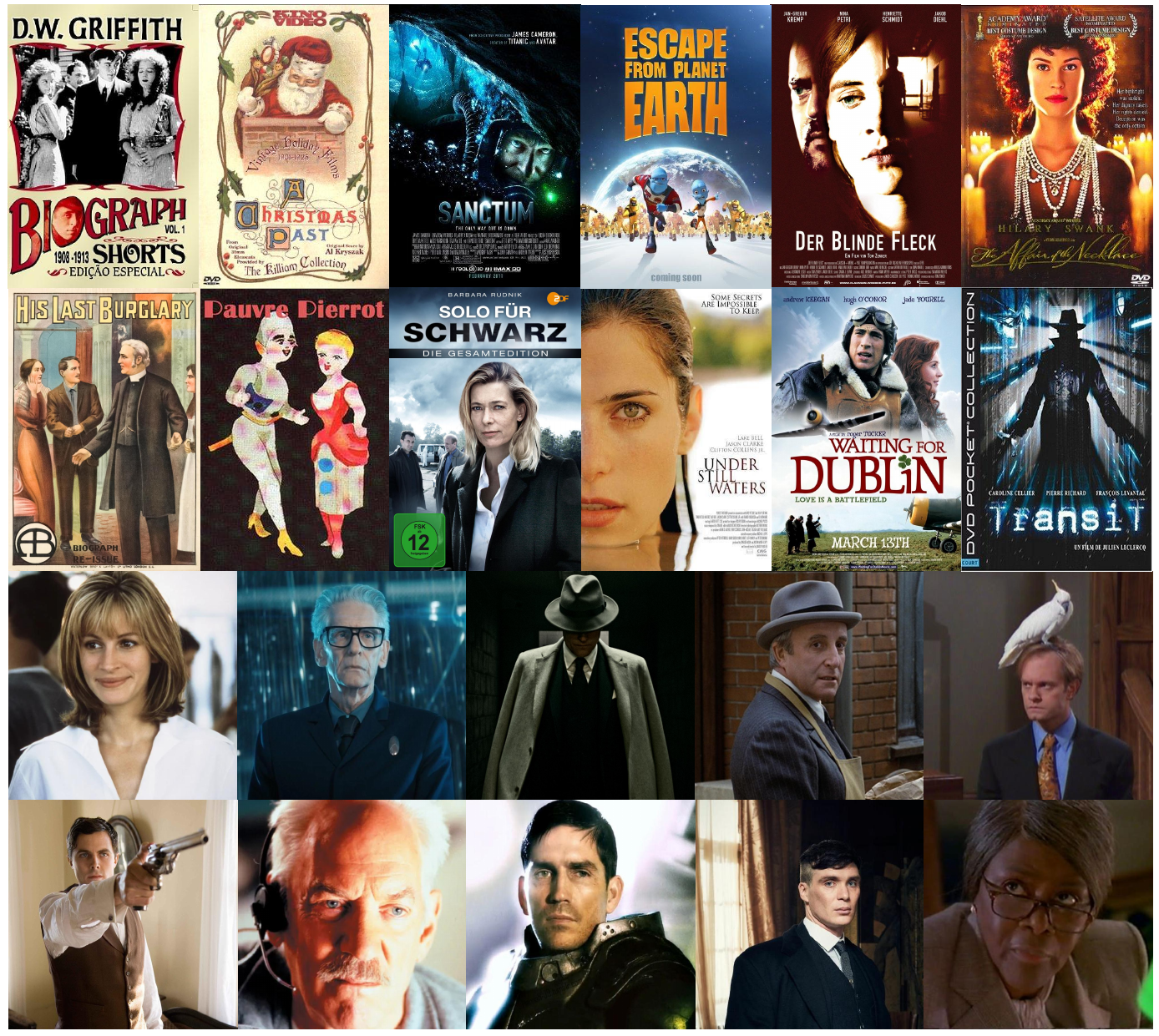}
	\caption{Examples of posters and actor images in MPDS.}
	\label{generation2}
\end{figure}

Thanks to IMDB's detailed categorization of movies, we were able to handle NSFW content effectively. We dropped movie posters for films classified as NSFW, such as those rated R, Adult, PG-18. When obtaining actor photos, we used a sequential strategy, which resulted in the indiscriminate collection of individuals related to the movie posters. This included not only actors but also behind-the-scenes personnel such as directors and screenwriters. This step was essential to streamline our dataset and maintain its relevance to the task. From the scraped pages, we extracted the image URLs to download the original images. Additionally, we filtered out any samples where the image size was smaller than 5KB, ensuring that our dataset only included high-quality images. The posters and their alt text cannot be used as image-text paired data because the alt text provides an abstract description of the posters. Training diffusion models requires images paired with their literal descriptions. Using only manually annotating hundreds of thousands of images is both expensive and inefficient, and it does not guarantee consistent annotations. Therefore, we used the Vision-Language Model Blip2  combined with manual annotation to annotate literal descriptions for the images. The literal descriptions are clearly more suitable as literal descriptions of the images compared to the original alt text. This approach ensures more accurate and relevant image-text pairs for our diffusion model training, enhancing the overall quality and effectiveness of the dataset. 

\subsection{Dataset Composition}
The dataset consists of two folders, which provide images about posters and actors, and three CSV file, meta.csv, posters.csv, and actors.csv, which provide important information about movies, posters' literal descriptions, and actors, respectively. The structure of each of these three CSV files is as follows:

\begin{itemize}
	
	\item \textbf{meta.csv}: This file contains data with key information related to the movies that describe movies from multiple aspects. Each row represents a movie. The attributes of this file are as follows:
		\begin{itemize}
			\item ‘id’ (int): A unique integer identifier for a movie, which is the primary component of the movie detail page URL. The identifiers start from the integer 1, but they are not consecutive because some URLs are not valid.
			\item ‘movie-url’ (str): The URL of the movie datails page. The URL is primarily associated with the id attribute. For example, for a movie with the id 2024, the URL format is ``https://www.imdb.com/title/tt0002024", which is composed of ``https://www.imdb.com/title/tt" followed by seven digits. For ids with fewer than seven digits, zeros are added in front to make up seven digits.
			\item ‘title’ (str): The movie title. Most of the movie titles are in English, we dropped the few that are not. 
			\item ‘genres’ (list): The movie genres. A movie usually belongs to multiple genres. This attribute contains one to three primary genres of the movie. Movies of the same genre generally have similar poster styles. Through this attribute, different types of movie posters can be selected for training.
			\item ‘actors’ (dict): The main cast of the movie. This attribute contains actor names as keys and their corresponding details web URLs as values.
			\item ‘poster-url’ (str): The URL of the poster. The URL for the high-resolution version of the poster, which can be used to download the poster.
			\item ‘synopsis’ (str): The synopsis of the movie.
 		\end{itemize}
	\item \textbf{posters.csv}: This file contains data related to the poster. Each row represents a movie poster and its literal description, which can used for text-to-image generation model. The attributes of this file are as follows:
	\begin{itemize}
		\item ‘id’ (int): A unique integer identifier for a movie poster. This id is the same as the one in meta.csv, and the downloaded posters in the folder are named using the id, such as ``2024.jpg".
		\item ‘text’ (string): A literal description of the image represented by the id.
	\end{itemize}
\begin{figure}
	\centering
	
    \includegraphics[width=1\linewidth]{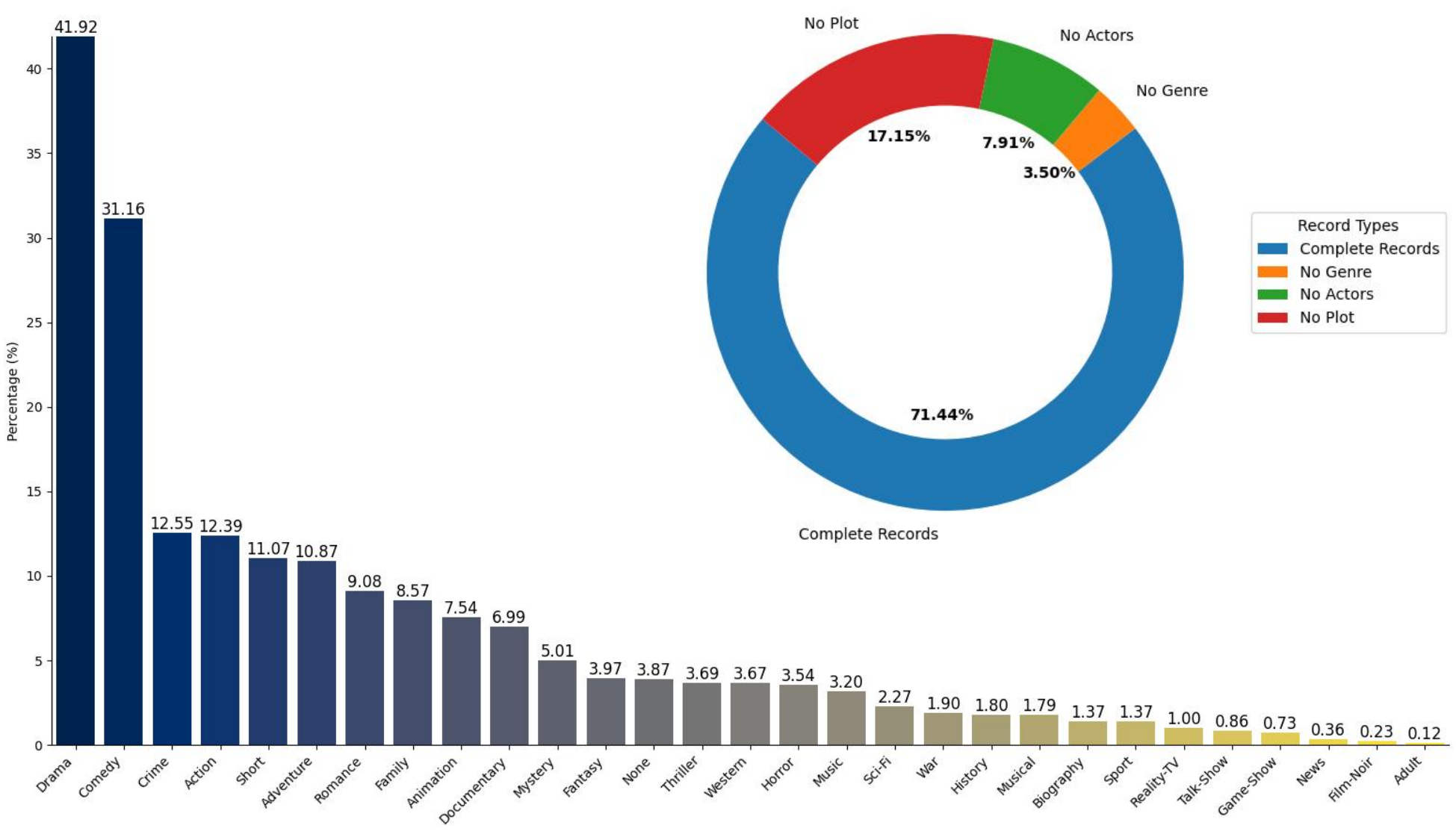}
    
	\caption{ Statistics of movie genres and movie attributes. Below is a bar chart showing the percentage of movie genres, above is a pie chart about recording types. }
  
	\label{statistic}
\end{figure}

	\item \textbf{actors.csv}: This file contains data related to the actors. Each row represents an actor and photos of the actor. This file is mainly used for the subject-driven text-to-image generation model. The attributes of this file are as follows:
	\begin{itemize}
		\item ‘id’ (int): A unique integer identifier for an actor, which is the primary component of the actor detail page URL.
		\item ‘actors-url’ (str): The URL of the actor details page. The URL is also associated with the id attribute. The URL format is ‘‘https://www.imdb.com/name/nm0002024". Its composition is similar to the movie-url in meta.csv.
		\item ‘actors name’ (str): The actor name.
		\item ‘image-url’ (str): The URL of the actor's images. Each actor has approximately one to two images.
	\end{itemize}
	
\end{itemize}
\subsection{Dataset Statistics}

\begin{figure*}[h]
	\centering
	\includegraphics[width=1\linewidth]{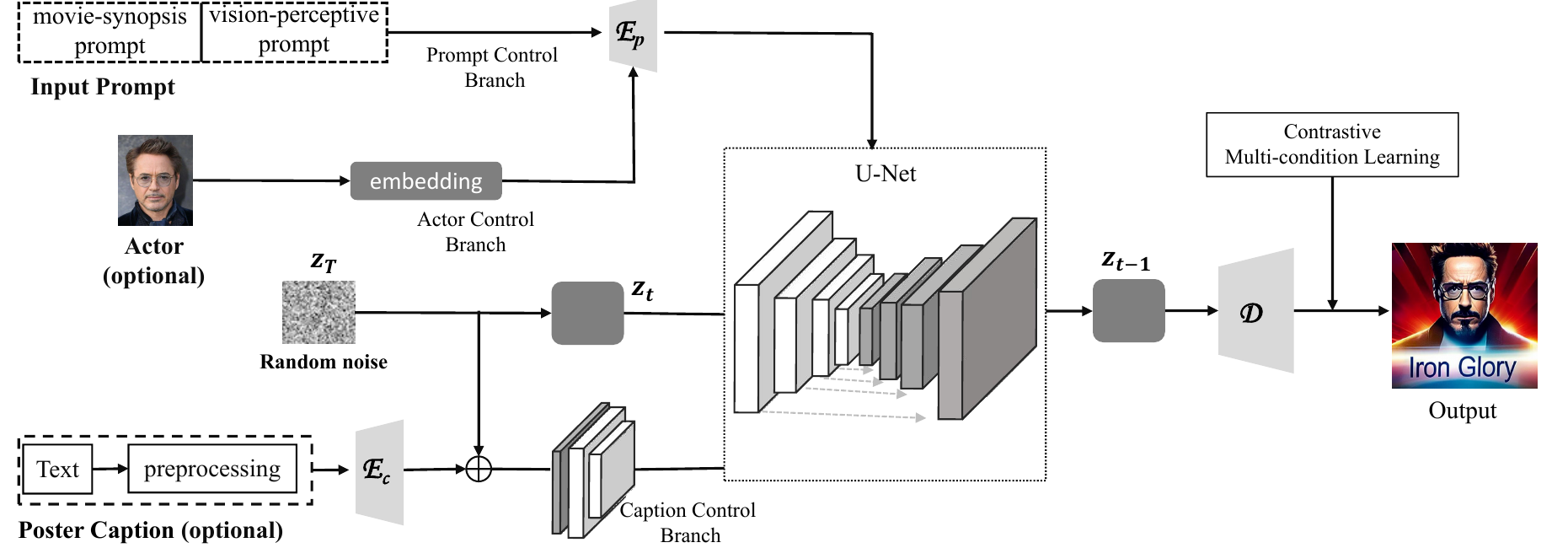}
	\caption{The proposed multi-condition diffusion framework for movie poster generation.Given the input prompt $\mathbf{I}_{prompt}$, the input poster caption $\mathbf{I}_{c}$, and the input facial image of an actor $\mathbf{I}_{a}$, the proposed framework can generate images in the stype of movie posters. Concretely, the generated posters encompass: i) the scenes that align with the prompts, which consist of two parts including the movie-synopsis prompt and vision-perceptive prompt (i.e. poster scene description); ii) the role played by the designated actor; iii) the texts in poster that match the specified content of the input caption. Specifically, the poster caption $\mathbf{I}_{c}$ and actor face $\mathbf{I}_{a}$ are optional to achieve personalized poster design.}
	\label{flow}
\end{figure*}

This dataset includes 373k image-text pairs, where the data are primarily from movie and TV show posters on IMDB. It contains key information about the movies, such as their genres. Each movie can be classified into one to three genres. There are a total of 28 different genres in the dataset, with their respective proportions shown in Fig.\ref{statistic}, where the top three are Drama, Comedy, and Crime. Drama and Comedy genres have the largest proportions, accounting for approximately 41.92\% and 31.16\% respectively. 

Our metadata is sourced from IMDB, a website known for its high information security. The presence of NSFW content, such as Adult-rated movies, is minimal, accounting for only 0.12\%. We have removed records of movies with these types, ensuring the overall security of our dataset.

Around 14k (3.50\%) of the movies do not have genre information. Furthermore, 31k (7.91\%) movie records lack actor information, and 67k (17.15\%) lack plot descriptions. Most of the movies lacking these attributes are older and more niche, but they all have the attribute of a poster-url. Therefore, we did not drop them.

\section{Experiments}
\label{propos}
\subsection{The proposed multi-condition diffusion framework}

As Fig.~\ref{flow} shows, the input prompt $\mathbf{I}_{prompt}$ passes through a text encoder and is then fed into the U-Net block as a text-level condition. Simultaneously, the optional poster caption undergoes preprocessing (including sketch render and edge detection), then passes through a ControlNet~\cite{zhang2023adding} to serve the caption control branch. Also, inspired by the textual inversion~\cite{gal2022image}, the actor images (if provided) are encoded into a specific embedding added to the text encoder, and is employed as the input of the actor control branch. Both the caption control and actor control branches serve as image-level conditions to guide the poster design. Under the guidance of the multiple conditions above, i.e. one text-level condition and two branches of image-level conditions, the U-Net block performs the forward and backward diffusion process, and finally reconstructs the output image from Gaussian noises. Moreover, we integrate the localized attention constraint presented in~\cite{zhang2024brush} to alleviate the risk of incorrect fusion among the text-level and two image-level controls. To further refine the role and caption generation in posters, a contrastive multi-condition learning process is performed to promote the naturalness of foreground-background blending. Please see the details in Appendix~\ref{Appendix}.

\subsection{Implementation Datails}
\subsubsection{Fine-tuning model with the MPDS}
The model is built with Diffusers. The pre-trained model used is 'runwayml/stable-diffusion-v1-5'. We fine-tune the model using the LoRA strategy, setting the base learning rate to 1e-4 and the dimension of the LoRA update matrices to 4. The training images and text come from our MPDS dataset, and the input images are maintained at a resolution of 512 × 512. Taking both training effectiveness and costs into consideration, we randomly sampled 2,000 examples for fine-tuning.
\subsubsection{Multi-condition diffusion model}
The pre-trained model used in our proposed Multi-condition diffusion framework is the model fine-tuned with the MPDS dataset, as previously mentioned. In the caption control branch, we mainly use the pre-trained models from 'lllyasviel/sd-controlnet-canny' to incorporate 512x512 canny images as conditional inputs into the model. In the Actor Control Branch, each actor is represented by 3-5 photos from the MPDS dataset, each with a resolution of 512x512, to extract features of the actor in the stills, which helps generate better personalized movie posters. We used one V6000 GPU with a batch size of 4. The base learning rate was set to 5e-4, and the minimum learning rate was set to 1e-6. The model was trained for 500 steps.

\subsection{Experimental Results}
\label{result}
\begin{figure*}[h]
	\centering
	\includegraphics[width=1\linewidth]{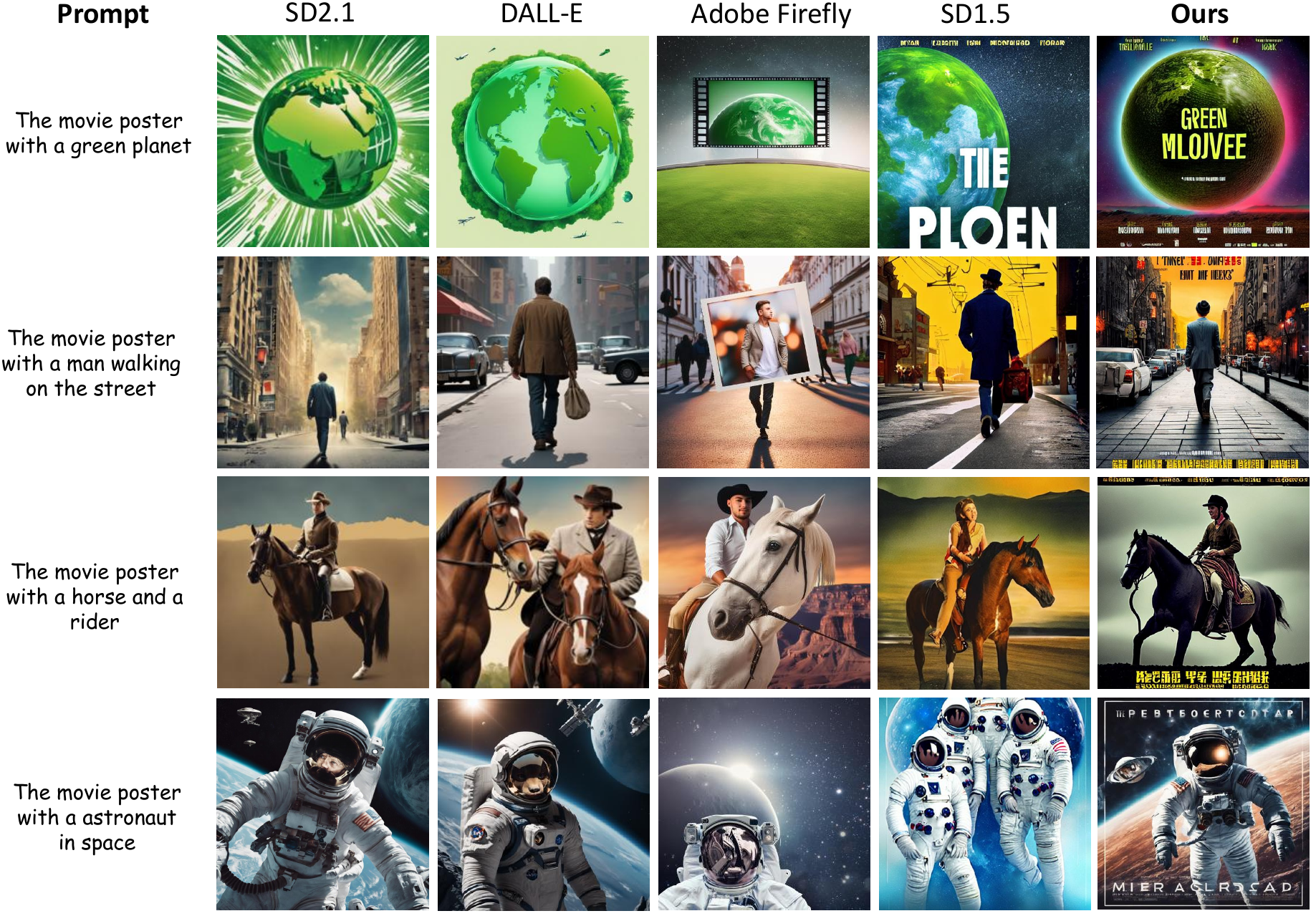}
	\caption{Visualizations of movie poster generation compared with existing methods, the corresponding textual prompt for the images below. }
	\label{generation5}
\end{figure*}

To evaluate the effectiveness of the constructed MPDS in promoting the movie poster generation, as well as the performance of our proposed framework, we conduct both quantitative and qualitative comparisons, and the experimental results are presented in Table~\ref{evaluation} and Fig.~\ref{generation5}, respectively.

\begin{table}[ht]
\centering
\begin{tabular}{lccc}
\toprule
Method & CLIP-score\(\uparrow\) & FID\(\downarrow\) & PickScore\(\uparrow\) \\
\midrule
SD1.5 & 0.8206 & 449 & 24.9\% \\
SD2.1 & 0.7976 & 436 & 42.7\% \\
DALL-E & 0.8145 & 504 & 48.9\% \\
Adobe Firefly & 0.7694 & 435 & 36.1\%\\
\textbf{Ours} & \textbf{0.8533} & \textbf{420} &\textbf{61.9\%} \\
\bottomrule
\hline
\end{tabular}
\caption{Quantitation comparison with existing methods. The bold numbers represent the best results among all compared methods.}
\label{evaluation}
\end{table}

In the quantitative comparison, we selected the following three metrics:
(i) \textbf {CLIPScore}~\cite{radford2021learning} is used to measure the similarity between the generated images and the input prompts. 
(ii) \textbf{FID}~\cite{heusel2017gans} used to measure the distribution distance between the generated movie poster set and the real movie poster set.
(iii) \textbf{PickScore}~\cite{kirstain2023pick} is a CLIP-based scoring function, which is mainly used for predicting human preferences to comprehensively assess the visual quality and text alignment of images.
We demonstrate the quantitative results compared with existing methods in Table~\ref{evaluation}. The comparison methods are SD1.5, SD2.1, DALL-E, and Adobe Firefly. To test their performances, we directly use those released models to generate movie posters based on the same prompts. As shown in Table~\ref{evaluation}, after training on the MPDS, our proposed model presents the state-of-art performances in generating movie posters by achieving a competitive PickScore and also attaining the best results in Clip-score and FID. 

Fig.~\ref{generation5} presents a qualitative comparison with various state-of-the-art text-to-image generation models. Compared to the baselines, our method exhibits a significant advantage in movie poster generation. Specifically, we give examples of four distinct prompts to evaluate the capacities of poster generation. It is demonstrated that following fine-tuning on the proposed MPDS dataset. As Fig.~\ref{generation5} shows, our method outperforms the existing methods in generating consistent and satisfactory results in terms of text, style, and layout. Taking a prompt example of "The movie poster with a green planet", Adobe Firefly, DALL-E, and SD2.1 struggle to generate text on the image, whereas SD1.5 synthesizes text which is contrary to the visual elements. A possible explanation is that these methods lack a specialized text processing module, as well as their lack of training on a specialized poster generation dataset. In comparison, our proposed approach, following refinement on the MPDS dataset, generates poster cation that harmoniously integrates with the thematic elements of the input poster, thereby yielding visually plausible results. The generated high-quality results verify the effectiveness of our proposed dataset in promoting the movie poster generation task.

\subsection{Ablation Studies}
\label{ablation}
\begin{figure*}[h]
	\centering
	\includegraphics[width=1\linewidth]{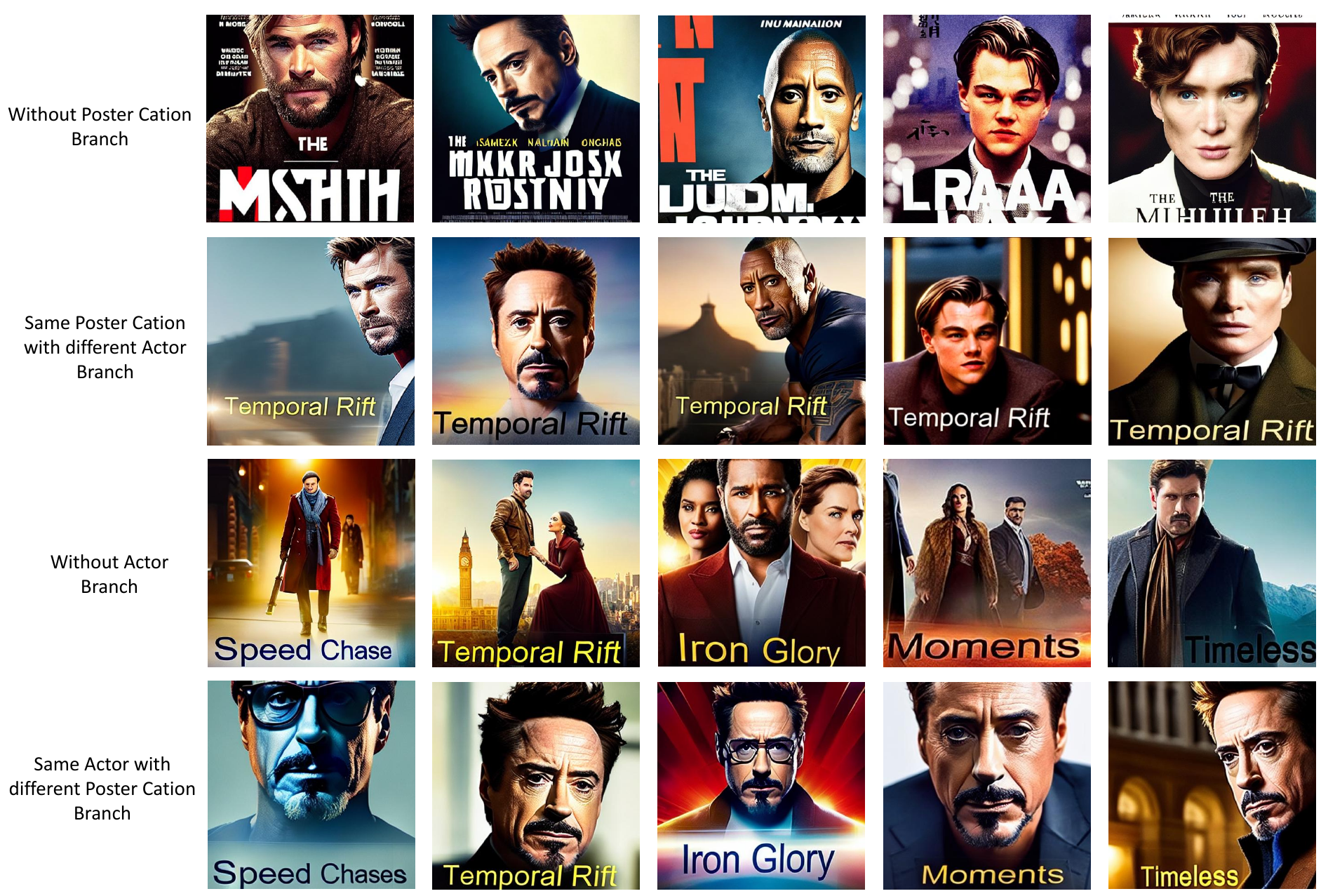}
	\caption{Ablation experiment results with/without different Control Branches.}
	\label{personal}
\end{figure*}

In this section, we conduct an extensive ablation study of the proposed framework based on different Control Branches. As shown in Fig.~\ref{personal}. In our study, we design experiments under the specific conditions of ‘‘Without Poster Caption Branch" and ‘‘Without Actor Branch", seeking to elucidate their pivotal roles in poster generation. The compelling findings underscore the critical significance of these two conditions in shaping the outcome of the generative process.  Furthermore, extending our investigation, we delve into experiments conducted within the parameters of ‘‘Same Poster Caption with different Actor Branch" and ‘‘Same Actor with different Poster Caption Branch", effectively reinforcing the superior efficacy of our proposed framework.

\section{Dicussion}
\label{dicussion}
\subsection{License}
The dataset is distributed under the Creative Commons 4.0 BY-NC-SA (Attribution-NonCommercialShareAlike) license. The choice of this license is motivated by the intended uses of the dataset and other considerations that we discuss next.

\subsection{Intended Uses}
The dataset proposed in this article will mainly be applied to poster generation, is intended to be used for academic research only. It can be divided into subsets based on movie genres to train models for generating posters of specific genre types. It can also be partitioned by actors, as the same actor may appear in dozens of movies. Through well-segmented text pairs for training, it is possible to generate movie poster styles similar to those in which a particular actor has appeared. The presentation of an individual actor in their personal photo and in a movie poster varies, so training with both types of images may yield different effects. 
\subsection{Limitations}
\label{limitations}
MPDB provides downloaded images and posters, requiring approximately 100GB of hard drive space, along with a CSV file containing the download links for the images. We used the Python requests library with ten terminals to batch download the images, which took about 10 hours. Due to occasional instability of IMDB's servers and some invalid URLs, a very small portion of images failed to download, which can be ignored. For users with special requirements for training data, we recommend using meta.csv and posters.csv to filter the necessary image URLs for manual download.

The information about a movie is complex. Although we have collected many attributes about movies, it is not comprehensive. The existing attributes include the URL of the movie's detail page, from which users can collect the information they need.

Additionally, the literal descriptions of the posters in MPDB were generated using the vision-language model Blip2. Although Blip2 is quite powerful, we recommend users to try different VLMs or other methods to process the images.

\subsection{Ethical Considerations}
In this study, we adhered to ethical guidelines throughout the collection and analysis of movie posters and related information. The meta data was obtained from IMDB~\cite{nichol2021glide}. There is no bias in the process of data collection and processing.

We respected all applicable copyright protections, ensuring that our use of movie posters and related information does not infringe upon the rights of the creators or copyright holders. The data was used solely for research purposes and cannot be used for commercial activities, aligning with the original intent of the data providers. We will inform users of this when providing the dataset.

\subsection{Future Work}

As of now, we have successfully collected and processed 373,000 data entries. Moving forward, we plan to continue gathering and processing new data, with the expectation that our dataset will eventually reach one million entries. Given that new movies are released daily, we will regularly collect and process the latest data to keep our database up to date. In addition, we will identify, filter, and analyze new information that is not yet included in our current dataset but may prove valuable. Based on the quality, availability, and feasibility of this new information, we will update and expand our dataset accordingly. This ongoing effort ensures that our data remains comprehensive, relevant, and useful for future analysis and applications.

\section{Conclusion}

The MPDS dataset demonstrated significant potential in advancing personalized movie poster generation. By integrating movie plots with visual perceptual cues, MPDS not only addresses the shortcomings of existing datasets but also provides a robust foundation for future image generation models. Our experimental results indicate that diffusion models trained on MPDS data outperform traditional models in terms of generation quality and controllability. Furthermore, the creation of MPDS opens up opportunities for further research and optimization, such as exploring the integration of multimodal data or incorporating more emotional text prompts. In the future, we plan to continue refining the dataset and expanding it to other types of visual generation tasks to further validate its broad applicability and robustness.
\newpage

\bibliography{aaai}

\appendix
\section{Appendix}
\label{Appendix}
For personalized and refined poster design, we coordinate the text-level condition $\mathbf{I}_p$ and two image-level conditions, i.e. the actor image $\mathbf{I}_a$ and poster caption image $\mathbf{I}_c$ through contrastive multi-condition learning. In the backward diffusion, we have the following formulation:

\begin{align}
z_{t-1}=\epsilon (z_t, \mathbf{I}_p, \mathbf{I}_a, \mathbf{I}_c).
\end{align}

According to the Bayes rule, we have:

\begin{align}
P(z_t| \mathbf{I}_p, \mathbf{I}_a, \mathbf{I}_c)= \frac{
P(\mathbf{I}_p|z_t)P(\mathbf{I}_a| \mathbf{I}_p, z_t)P(\mathbf{I}_c| \mathbf{I}_p,\mathbf{I}_a, z_t)P(z_t)}{P(\mathbf{I}_p, \mathbf{I}_a, \mathbf{I}_c)}.
\end{align}

By applying the logarithm operation on both sides of the equation and further taking differentiating. we can derive:

\begin{align}
\nabla_{z_t}P(z_t| \mathbf{I}_p, \mathbf{I}_a, \mathbf{I}_c)  & = \nabla_{z_t} P(z_t) + \nonumber \nabla_{z_t} P(\mathbf{I}_p|z_t) \\
&+ \nabla_{z_t} P(\mathbf{I}_a| \mathbf{I}_p, z_t) + \nabla_{z_t}P(\mathbf{I}_c| \mathbf{I}_p,\mathbf{I}_a, z_t). 
\end{align}

The above equation can be also represented by:

\begin{align}
\nonumber  z_{t-1} & = \tilde \epsilon (z_t, \mathbf{I}_p, \mathbf{I}_a, \mathbf{I}_c) \\
\nonumber          & = \epsilon (z_t, \varnothing, \varnothing, \varnothing) + s_1 (\epsilon(z_t,\mathbf{I}_p, \varnothing, \varnothing)-\epsilon (z_t, \varnothing, \varnothing, \varnothing)) \\
\nonumber          & + s_2 (\epsilon(z_t,\mathbf{I}_p, \mathbf{I}_a, \varnothing)-\epsilon (z_t, \mathbf{I}_p, \varnothing, \varnothing)) \\
\nonumber          & + s_3 (\epsilon(z_t,\mathbf{I}_p, \mathbf{I}_a, \mathbf{I}_c)-\epsilon (z_t, \mathbf{I}_p, \mathbf{I}_a, \varnothing)).\\
\end{align}

\end{document}